# Mining Truck Platooning Patterns Through Massive Trajectory Data

Xiaolei Ma, Enze Huo, Haiyang Yu, and Honghai Li

*Abstract*—Truck platooning refers to a series of trucks driving in close proximity via communication technologies, and it is considered one of the most implementable systems of connected and automated vehicles, bringing huge energy savings and safety improvements. Properly planning platoons and evaluating the potential of truck platooning are crucial to trucking companies and transportation authorities. This study proposes a series of data mining approaches to learn spontaneous truck platooning patterns from massive trajectories. An enhanced map matching algorithm is developed to identify truck headings by using digital map data, followed by an adaptive spatial clustering algorithm to detect trucks' instantaneous co-moving sets. These sets are then aggregated to find the network-wide maximum platoon duration and size through frequent itemset mining for computational efficiency. We leverage real GPS data collected from truck fleeting systems in Liaoning Province, China, to evaluate platooning performance and successfully extract spatiotemporal platooning patterns. Results show that approximately 36% spontaneous truck platoons can be coordinated by speed adjustment without changing routes and schedules. The average platooning distance and duration ratios for these platooned trucks are 9.6% and 9.9%, respectively, leading to a 2.8% reduction in total fuel consumption. We also distinguish the optimal platooning periods and space headways for national freeways and trunk roads, and prioritize the road segments with high possibilities of truck platooning. The derived results are reproducible, providing useful policy implications and operational strategies for large-scale truck platoon planning and roadside infrastructure construction.

## I. INTRODUCTION

THE freight sector accounts for 60% of cargoes, and this trend continues to increase rapidly [1]. However, long-distance and high-volume logistics operations consume a large amount of fuel. In the European Union, the energy consumption of road freight transportation is 27% among all transport modes, leading to 20% carbon emission of total emissions [2]. Truck platooning is considered one of the most promising cutting-edge technologies to alleviate energy consumption and environmental pollution, and refers to a group of trucks traveling with relative short headways for a long period via telecommunication [3], [4]. Truck platooning can effectively reduce aerodynamic resistance between leading and following vehicles, and is thus able to lower the total energy consumptions of trucks [5]. Truck platooning can also improve road safety and enhance capacity aside from offering energy savings [6], [7], [8]. With the abovementioned features, truck platooning is undoubtedly a hot topic. Theoretically speaking, truck platooning can be a leader-following consensus problem. In the past decades, a myriad of control theory-based approaches are proposed. Zhang et al. [9] considered a leader-follower consensus of multiagent system with communication capability and energy constraints, and modeled the system using Markovian approach and Lyapunov stability theory. Zhang et al. [10] investigated the issue of leader-following consensus in a linear and Lipschitz nonlinear multiagent system with limited information. Wen et al. [11] proposed a neural network-based leader-follower consensus scheme. Tan et al. [12] developed a distributed event-triggered impulsive control method to study the leader-following consensus issue. You et al. [13] studies the leader-following consensus problem in the environment of high-order nonlinear multiagent system, and used a self-trigger and dynamic output feedback control scheme. Yue and Meng [14] extended the leader-following consensus issue into a cooperative set problem, where multiple agents can be aggregated into a desired set. They proposed an approximate projection algorithm and proved its convergence conditions. Liu et al. [15] emphasized the synchronization of mobile agent platoon, where each agent's speed and heading is different. Petrillo et al. [16] further designed a control mechanism to resist cyberattacks in a connected and autonomous vehicle (CAV) platoon. The above solid theoretical work sheds substantial light on cybernetic principles of truck platooning but still focuses on simulation rather than empirical validation.

Bhoopalam et al. [17] divided truck platooning into three categories: scheduled platooning, real-time platooning, and opportunistic platooning. The former two tend to be applied to regional planning with prior intervention and coordination, and are also known as planned platooning, whereas the latter more likely happens in a realistic environment where trucks that are in close proximity are ad-hoc informed about platooning. For planned platooning, logistics schedules, origin/destination

This paper is supported by the National Natural Science Foundation of China (52072017). (*Corresponding author: Xiaolei Ma*)

Xiaolei Ma, Enze Huo, Haiyang Yu and Honghai Li are with the Beijing Key Laboratory for Cooperative Vehicle Infrastructure System and Safety Control, School of Transportation Science and Engineering, Beihang University, Beijing 100191, China, and also with the Beijing Advanced Innovation Center for Big Data and Brain Computing, Beihang University, Beijing 100191, China (e-mail: xiaolei@buaa.edu.cn; chirshuo@buaa.edu.cn; hyyu@buaa.edu.cn; honghai_1@126.com).

information (i.e., locations and times), and delay tolerance are often acquired in advance to minimize network-level fuel consumption or maximize the platoon size. However, industrial resistance and practical factors hinder the wide implementation of planned platooning. Logistics operators are reluctant to change their route delivery schedules and routes to cater to platooning with other companies [18]. Compared with passenger vehicles, trucks' estimated arrival time are more uncertain due to road congestion, overload inspection, and driving hour limitation, making network-wide truck platooning optimization impossible to solve in a finite time frame as a NP-hard issue.

Fortunately, high-fidelity truck GPS data from freight fleet management systems reproduce trucks' actual spatiotemporal footprints and can be utilized to coordinate trucks to form platoons by identifying leaders and followers from massive trajectories. Such adjustments do not require prior planning, and truck platoons can be spontaneously formed by slightly adjusting speeds on certain road segments rather than changing routes. As indicated by Liang et al. [19], the energy savings induced by truck platooning can be counteracted by the energy consumptions by detouring to seek platooning opportunities. In addition, as the national road network density increases, investing in and constructing roadside units on each segment is not realistic to facilitate V2V communications. Highlighting road segments with high platooning possibilities will be beneficial to prioritize the existing roadside facilities to be upgraded for truck platooning.

Therefore, this study aims to develop framework to mine massive truck trajectories to identify possible truck platoon patterns. The proposed trajectory mining procedure is to detect leaders and followers within a short distance for a given time period. The identified patterns actually reduce the computational complexity of truck platoon planning by emphasizing the speed coordination between leading and following trucks in a large-scale network. The entire framework is composed of four steps. In the first step, a modified map matching algorithm considering relative driving direction is proposed to map each truck's locations on the OpenStreetMap (OSM) network. We further recognize the instantaneous co-driving sets with the same directions and close spacing by using enhanced ordering points to identify the clustering structure (OPTICS) algorithm. Among the generated sets, the spontaneous truck platooning patterns are found based on frequent itemset mining in depth-first spanning trees. In the final step, a fuel consumption estimation model is adopted to quantify the energy savings of the identified truck platoons, where all parameters are calibrated from either field data or existing literature. We systematically evaluate the platooning potentials and performance by using several metrics. To summarize, the primary contributions of this study are twofold:

(1) A series of trajectory mining approaches is proposed to identify spontaneous truck platooning patterns. To be specific, an enhanced OPTICS algorithm is proposed to detect instantaneous co-driving sets from millions of truck GPS points. This algorithm replaces the original Euclidean distance with the following distance between trucks and considers driving direction, roadway junction, and angle of reachability distance sequence, thus improving the detection rate of sub-centralized and speed-coordinated truck fleets. The largest truck set with the longest co-driving time-period is extracted from the instantaneous co-driving sets of the entire day by using frequent itemset mining algorithm.

(2) On the basis of massive truck trajectory data in Liaoning Province, China, we answer how many trucks could be coupled as platoons and how much fuel savings could be achieved for a specific day. We also prioritize those road segments with high platooning potential, which can provide solid policy support to upgrade existing roadside facilities for automating freight transportation.

The remaining part of this paper is outlined as follows. Section 2 reviews different truck platooning modes and emphasizes trajectory mining approaches for co-movement pattern identification. Section 3 describes the methodological framework, which involves a modified hidden Markov model (HMM)-based map matching algorithm tailored for truck platooning, an enhanced OPTICS algorithm to recognize instantaneous co-driving sets, and a pruning algorithm to find the potential instantaneous platooning patterns. We further introduce the fuel consumption estimation model. Section 4 provides the detailed data collection efforts and several platooning performance indicators. Section 5 analyzes one-day truck trajectory data collected in Liaoning Province, China, and evaluates the platooning performance and energy savings by using the proposed spontaneous platooning pattern mining approaches. The spatiotemporal distribution of truck platoons is visualized in a map. Section 6 presents the conclusions and envisions the future directions.

## II. LITERATURE REVIEW

Depending on whether trip planning is involved, the form of truck platooning can be divided into scheduled platoon planning and unscheduled platoon planning [17] On the basis of the timeliness of trip announcements, the scheduled platoon planning can be further divided into static platoon planning and real-time platoon planning. The unscheduled platoon planning is also referred to as opportunistic platooning or spontaneous platooning, which is the emphasis of this study. In the subsequent literature review section, we systematically discuss the methodological advances of both scheduled and unscheduled platoon planning.

For scheduled static platoon planning, truckers can decide whether to deviate from their planned routes for platooning. Moreover, many practical restrictions are in place for forming platoons. According to whether preplanned routes are allowable, the scheduled platoon planning methods can be divided into two categories. For platooning with fixed routes, Van de Hoef [20] determined the possible trucks that can adjust their speeds for platooning by ensuring on-time arrivals. The optimal speed profiles are given for each truck. Zhang et al. [21] consider travel time uncertainty and analyzes the two scenarios of two trucks traveling either to the same headings or meeting and diverging. Liang et al. [22] discussed the specific speed coordination approach for two trucks on the same route.

Farokhi and Johansson [23] investigated the influence of truck platooning on congestion pricing and speed of non-platooning trucks, and discussed the potential of platooning multiple trucks on a single route. Meisen et al. [24] introduced sequential mining methods to find the maximum overlapping routes on the basis of each truck's fixed path. Without the late-arrival issue being taken into consideration, truck platoon arrangements are proposed based on the length of overlapping segments, waiting time cost, and fuel savings, which dramatically improves the optimization efficiency. For dynamic route problems, where trucks can change their original routes for platooning, Larson et al. [25] introduced the concept of controllers located at various road junctions. The local controllers can perceive the position, speed, and OD information of incoming trucks for platooning by speed coordination. Larsson et al. [26] modeled the platoon optimization as a graph routing problem and introduced a subheuristic approach to solve. Larson et al. [27] and Nourmohammadzadeh and Hartmann [28] used mixed-integer planning and genetic algorithm to form platoons of 20–25 trucks at the state level. However, due to the limitation of the NP-hard problem, the size of optimized trucks is relatively small, which cannot be implemented in a large-scale freight network for real-time coordination. The uncertainties associated with overloading enforcement inspection and temporary maintenance further bring additional challenges to truck platoon planning issue [18].

For real-time platoon planning, dynamic programming is a common approach in most studies. The entire optimization process could be repeated when a new event is detected; for example, new freight plans are adjusted or trucks miss the platooning opportunities at one point [29], [20], [25]. Adler et al. [30] imagined a situation in which multiple trucks arrive at a particular station with Poisson distribution and are platooned to the same destination. Two platooning policies are compared: trucks are grouped to leave at a certain timestamp or when the predefined platoon size is reached. Results show that regulating the platoon size saves more energy than regulating the time period. It can be predicted that it will take many years or even a decade for the academia to comprehensively organize and optimize real-time truck platooning with real-time information.

For unscheduled platoon planning, the purpose is to identify a collection of trucks that meet by chance and spontaneously travel within a short space headway for a given time period. Logistics operators in Portugal have conducted a series of interviews with truck drivers, confirming that on roadways, especially highways, many groups of unknown trucks accidentally meet and travel together for a long distance until they are separated due to speed or route difference [18]. The key to a successful unscheduled truck platooning strategy is to determine instantaneous co-driving sets. To extract spontaneous platooning patterns, existing studies focus on iterative approaches that identify the platoon form-up times and durations. For the recognition of instantaneous co-driving sets, mainly three research directions exist. The first research effort utilizes density-based clustering approaches such as density-based spatial clustering of applications with noise (DBSCAN), mainly represented by Jeung et al. [31]. The second research area is flock search area identification. Liang et al. [32] regarded coupled trucks within a 100 m flock of the current truck and traveling at the same segment as the instantaneous co-driving sets, while Larson et al. [33] applied a search radius ranging from 0.5 to 5 km to estimate the coordination potentials for spontaneous truck platooning in a transportation network in Germany. On the basis of Liang et al. [32], Shein et al. [34] combined flock sets with a central distance less than the search radius at each timestamp. Therefore, the enhanced algorithm mitigates the insufficient adaptability of fixed radius and avoids excessive density connection caused by density clustering algorithms. The idea solution should comprehensively consider the heading and distance difference between trucks for flock recognition such as Andersson et al. [35].

The abovementioned approaches cannot effectively adapt to the realistic road topology. The radius search methods are unable to identify truck grouping behavior on road junctions and parallel segments, where trucks travel in relatively short lateral distances. Mistakenly clustering those trucks in different road segments as the same platoon through fixed radius searching is inevitable. The majority of iterative approaches apply the fixed radius to find co-driving trucks and cannot meet the address the issue of dynamic spacings between trucks. Shein et al. [34] alleviated this issue to some extent but still cannot eliminate the errors of choosing the initial trucks. In addition, almost all bidirectional trunk roads are represented as single lines in OSM. Thus, the approach proposed by Liang et al. [32] has a high possibility of recognizing trucks driving in opposite directions as instantaneous co-driving sets.

The current research on truck platooning does not consider freight network design to reap the maximum benefits of platooning. A platoon requires the support of roadside units; thus, logistics operators can better plan communication facilities by visualizing spontaneous truck platoon patterns and by highlighting the segments with high platooning potentials [17].

The focus of this study is spontaneous truck platoon planning based on massive trajectory data. We replace the geographical distance with the vehicle following distance to address the single-line representation issue in a digital map. An adaptive clustering algorithm to further refine instantaneous co-driving sets is proposed to alleviate excessive density connections. Truck platoon patterns are extracted based on instantaneous co-driving sets by using frequent itemset mining approaches. Finally, a series of policy recommendations is made to prioritize roadside communication infrastructures to enhance the effectiveness of truck platooning with a minimum transportation facility investment.

### III. METHODOLOGY

We consider a road network where multiple trucks travel with latitudes, longitudes, and timestamps. The problem is to find how many trucks can be coordinated with each other and grouped as platoons as time evolves. This problem is similar to the issue of companion or flock discovery, where a companion or flock refers to a group of moving objects (e.g., animals or pedestrians) traveling together sporadically. Unlike animals or people, who move in space and time flexibly, truckers have more strict schedules to follow and more fixed routes to travel. This problem becomes very complicated considering hundreds of millions of points concurrently scattering on a large-scale network with both spatial and temporal constraints. We aim to tackle this issue by using a series of data mining approaches.

In the following, a computational framework is presented to find the possible truck platoons by using massive trajectory data and digital maps. The framework is composed of three key steps. In the first step, we map each truck's locations on the corresponding road segments and also infer the truck's heading and following distance with the leading truck. We then develop a density-based clustering algorithm to detect the co-driving set at each timestamp. This algorithm can effectively group those trucks traveling on China's national expressways, which are commonly in parallel and hard to distinguish. The final step is to aggregate all sets for multiple time steps and find the most typical platoon patterns from numerous combinations. We applied the depth-first spanning trees with adequate pruning rules to lower the computational burden. The mined potential platoons are used to estimate the fuel consumptions.

*A. Map Matching with Heading*

The map matching procedure aims to accurately link each truck's latitude and longitude with each road segment given the high positioning errors of GPS devices. The highway attributes in China can be categorized as trunk roads and national-level expressways. The truck speed limits for national-level expressways and trunk roads are 100 and 60 km/h, respectively. Trucks occasionally travel on the trunk roads to avoid tolls on expressways. In most cases, trunk roads and expressways are in parallel and intersect with each other. In addition, trunk roads are represented by single lines in digital maps, thus incurring difficulties in separating opposite traffic by relying on truck trajectories only. Thus, distinguishing whether a truck travels on trunk roads or expressways with correct direction is not an easy task. In addition, to identify co-movement patterns, we are more concerned with the interaction between multiple truck trajectories rather than locating a single truck trajectory. Therefore, it is necessary to identify the relative driving direction of each truck to the matched segment at each moment and determine the following distance among multiple trucks.

Newson and Krumm [36] introduced the hidden Markov process into a map matching algorithm, where Lou et al. [37] pointed out that the transition probability should also consider the difference between speed calculated from GPS data and speed calculated from matched points. The relationship between positioning points and their corresponding segments can be regarded as the observations to the states, while the observation probability is determined by the vertical distance between the positioning points and the candidate segments within the given search radius. Therefore, determining the corresponding matched points and segments can be solved by dynamic programming influenced by both spatial and temporal factors. In OSM, each roadway is composed of segments with both start and end nodes. The default road direction is determined from the start node to the end node. However, most trunk roads allow trucks to travel on both directions but are presented using single lines. Moreover, these segments are not regularly connected from end to end. As a result, the default road direction is not fixed along each trunk road. The traditional HMM-based map matching algorithm cannot distinguish the driving direction relative to the matched segment direction. To improve the accuracy of instantaneous co-driving sets, each truck's heading on its corresponding segment needs to be identified. To address this issue, we extend the traditional HMM-based algorithm by incorporating four scenarios, where each truck's heading and segment direction is intertwined in Figure 1.

In Figure 1, $z_{t-1}$ and $z_t$ represent the truck's positioning points at timestamp $t-1$ and $t$, respectively. $s_{t-1}$ and $s_t$ indicate the matched segments of $z_{t-1}$ and $z_t$, respectively. $p(s_{t-1}, z_{t-1})$ and $p(s_t, z_t)$ are the two candidate points on both $s_{t-1}$ and $s_t$. A dashed line connects two segments to represent the virtual shortest network route. We aim to identify the heading of each truck GPS point relative to the matched segment direction. $dir = 0$ implies that the truck travels along the segment direction, while $dir = -1$ implies that the truck is away from the segment direction. The four scenarios are demonstrated in Figure 1, where the red line indicates the truck's actual heading, and $ToNode$ and $FromNode$ represent the nodes of matched segments that the truck is heading to and from, respectively. We iteratively calculate the transition probability of each scenario and select that scenario with the maximum probability, where the truck's heading relative to the corresponding matched segment at each timestamp can be determined. Notably, the low data quality of GPS points severely deteriorates the accuracy of map matching algorithm. To avoid the influence of low-quality positioning points on transition probability calculation, malfunctional data can be skipped, and previous truck locations with multiple timestamps ahead can be utilized as the reference data to jointly determine the truck's driving direction.

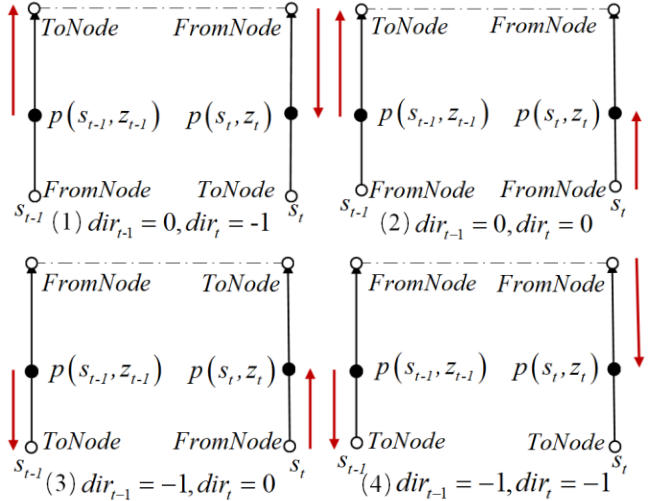

Fig. 1. Relationship between truck heading and segment direction.

*B. Instantaneous Co-driving Set Generation*

On the basis of the identified truck headings in the map matching process, we can further extract the set of instantaneous co-driving sets in the freight network at each moment. The so-called instantaneous co-driving set refers to a set of trucks with the same headings closely driving with each other and are distributed densely on road segments at certain time. However, the characteristics of transportation network, such as parallel-line construction, complex intersection, and lack of double-line description, bring challenges to truck co-driving set mining. Data in Liaoning Province in China extracted from OSM are taken as an example. Here, the shortest distance between national freeway G202 and trunk road G15 is only 30 m, and these two routes continue to expand in parallel. Thus, the conventional fixed-radius searching approaches cannot well recognize co-movement trucks traveling on the same route.

As an improved version of density-based clustering approaches, the OPTICS algorithm adapts to a changeable distance between individual points as the DBSCAN algorithm does. The dataset is ordered by keeping cluster hierarchy in a given parameter pool. Unlike the DBSCAN algorithm, which directly provides the clustering results of a dataset, the OPTICS algorithm first outputs the order in which the objects are processed and the information such as core distance and reachability distance. Given the requirement for a point $p$ to be a core point is there are $MinPts$ points in its $\varepsilon$-length flock, and the core distance $cd(p)$ defines the minimum radius to classify it as a core point. For a core point $p$, the reachability distance $rd(p,q)$ refers to the maximum between the core distance $cd(p)$ and the distance between two points. However, the OPTICS algorithm cannot be directly utilized to recognize trucks' instantaneous co-driving set because the original Euclidean distance is unidirectional. Under the circumstance of detecting multiple trucks' co-movements, each truck's direction needs to be captured. Hence, we introduce the concept of following distance in the OPTICS algorithm, namely, adaptive OPTICS (A-OPTICS). To simplify the problem, we extract the location of N on-line trucks at timestamp $t$ while using $o_p$ to represent their matched locations. Considering two trucks' locations $o_1$ and $o_2$ at timestamp $t$, the lengths of their matched segments are $Seglen_{o_1}$ and $Seglen_{o_1}$, respectively; the truck headings relative to the segment directions are $dir_{o_1}$ and $dir_{o_2}$, respectively. $r_{o_1}$ or $r_{o_2}$ represent the location of the matched point on each segment as a ratio of total segment length. On the basis of the results of map matching algorithms in Section 3.1, the start and end nodes of two matched segments can be identified as $FromNode_{o_1}, ToNode_{o_1}$, and $FromNode_{o_2}, ToNode_{o_2}$, respectively. Therefore, the following distance $FD(o_1, o_2)$ can be calculated as shown in Appendix 1, representing the geographic distance between following and leading trucks considering their relative headings. The co-driving relation is mutual; thus, the following distance $FD(o_1, o_2)$ is symmetric, that is $FD(o_1, o_2) = FD(o_2, o_1)$.

Therefore, the two-dimensional Euclidean distance in the original OPTICS algorithms is replaced by the one-dimensional following distance $FD(o_1, o_2)$, which can better depict the co-driving relationship among multiple trucks. If there are more than $MinPts$ trucks with the following distances to a truck lower than $\varepsilon$ kilometers, then this truck is considered a core object. The neighborhood distance $\varepsilon$ is set to 1 km to allow sufficient space for speed coordination and $MinPts$ should be set as 2 to form up a platoon. Therefore, for any given truck $o_p$, the core distance refers to the minimum following distance that makes the truck a core object within the given parameter $\varepsilon$ and $MinPts$, as shown in Equation 1.

$$cd(o_p) = \begin{cases} UNDIFINED, & if\ |N_\varepsilon(o_p)| < MinPts \\ FD\left(o_p, N_\varepsilon^{MinPts}(o_p)\right), & otherwise \end{cases} \quad (1)$$

where $N_\varepsilon(o_p)$ refers to those trucks whose following distances to $o_p$ are lower than $\varepsilon$, and $N_\varepsilon^{MinPts}(o_p)$ refers to the truck with following distance from $o_p$ to the $\varepsilon$-nearest trucks, while the core distance $cd(o_p)$ will satisfy $cd(o_p) \leq \varepsilon$ when $o_p$ becomes a core object. For any other truck $o_q$, the reachability distance of $o_q$ from $o_p$ is defined in Equation 2 under the given parameter $\varepsilon$ and $MinPts$. According to the OPTICS algorithm, the orders in which the trucks are processed at each timestamp are obtained, while the calculated reachability distance is denoted as $o_i.R$.

$$rd(o_q, o_p) = \begin{cases} UNDIFINED, & if\ |N_\varepsilon(o_p)| < MinPts \\ max\left(cd(o_p), FD(o_p, o_q)\right), & otherwise \end{cases} \quad (2)$$

The enhanced algorithm is still based on density clustering; therefore, the desired co-driving truck sets, which are the trucks that are densely connected with the core objects but are visibly far away with each other, could still be recognized as the same set. Figure 2 visualizes the uneven distribution of trucks' instantaneous locations on a single-line roadway. With the introduction of vehicle following and reachability distance, the initial set of the instantaneous co-driving set can be successfully found, as denoted by a dashed red circle. However, within the circle, $o_2$ and $o_8$ are sparsely distributed and should not be classified as a platoon with the remaining trucks.

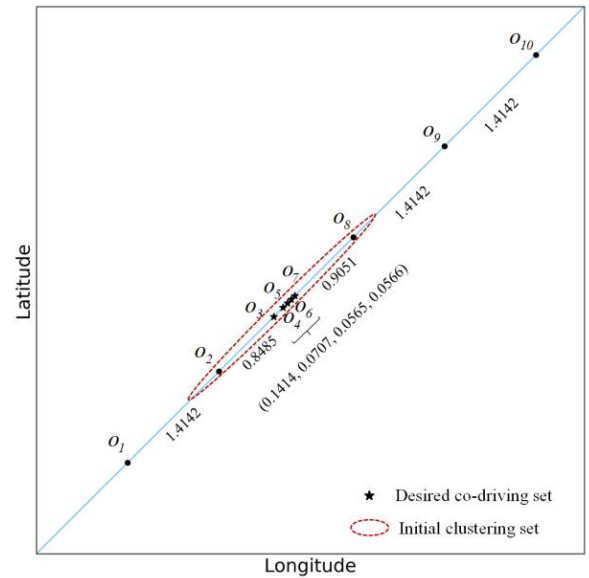

Fig. 2. Example of wrongly detected co-driving truck sets.

Fortunately, the variation tendency and reachability distance rate according to the clustering order can address this issue [38]. In Figure 3, to quantify the variation trend of reachability distance, the gap between any two consecutive trucks in the ordering is quantified as $\Delta$, two vectors $\overrightarrow{o_y o_x}$ and $\overrightarrow{o_y o_z}$ are constructed for any given truck $y$ and its two neighborhood trucks $x$ and $z$, respectively. The angle $\theta_{y.R}$ describes the variation tendency of reachability distance at the truck $y$, while the rate indicator $\Lambda_{y.R}$ is combined with $\theta_{y.R}$ to determine the boundary of the desired set in the ordering and can be computed

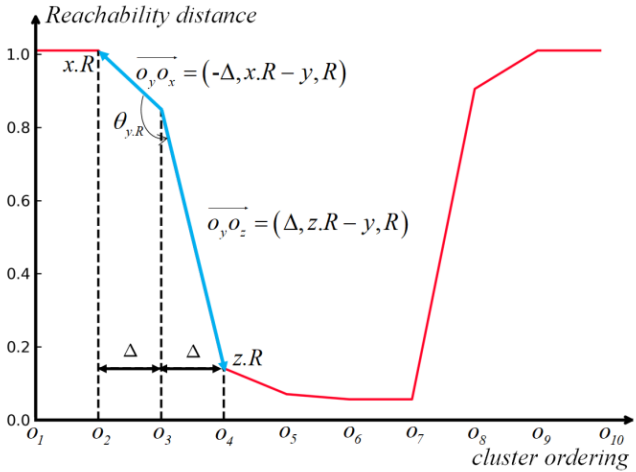

Fig. 3. The reachability distance diagram in OPTICS algorithm.

by Equations 3 and 4.

$$\theta_{y.R} = \langle \overrightarrow{o_y o_x}, \overrightarrow{o_y o_z} \rangle = arccos \left( \frac{-\Delta^2 + (x.R - y.R)(z.R - y.R)}{\|\overrightarrow{o_y o_x}\| \cdot \|\overrightarrow{o_y o_z}\|} \right) \quad (3)$$

$$\Lambda_{y.R} = \left| \frac{-\Delta}{x.R - y.R} \quad \frac{\Delta}{z.R - y.R} \right| \quad (4)$$

In simple terms, a smaller $\theta_{y.R}$ indicates that the truck $y$ is located toward the edge of a centralized desired subset (i.e., truck platoon), and $\Lambda_{y.R}$ can further assist in determining the initial and last trucks within the desired subset. Gap parameter $\Delta$ and the threshold values of $\theta_{y.R}$ and $\Lambda_{y.R}$ can be properly set based on the parameter $\varepsilon$ and $MinPts$. Given that the neighborhood distance $\varepsilon$ is set as 1 km, the minimum following trucks $MinPts$ is set as 2, the optimal gap parameter $\Delta$ should be 0.5, and the thresholds of $\theta_{y.R}$ and $\Lambda_{y.R}$ are 150° and 0, respectively. Moreover, the reachability distance threshold is 1, so that a truck that is neither outlier nor a core object will be 1.01. All the above parameters are well calibrated by extensive sensitivity analysis. We compute the variation tendency $\theta_{y.R}$ and reachability distance rate $\Lambda_{y.R}$ to improve the truck co-driving set in Figure 2.

A truck with $\theta_{y.R} < 150°$ and $\Lambda_{y.R} > 0$ is either the front of a platoon or an outlier nearest to the platoon. A truck with $\theta_{y.R} < 150°$ and $\Lambda_{y.R} < 0$ is either the end of a platoon or the second truck in the platoon. On the basis of the two rules, we can extract the first and last trucks in a centralized set based on the initial OPTICS clustering results. Detailed pseudocodes are provided in Appendix II.

### C. Spontaneous Platoon Pattern Mining

For a spontaneous platoon pattern $P(O, T)$, all trucks in all set $O$ are located in the same instantaneous co-driving set within the timestamp set $T$. Therefore, spontaneous platoon pattern mining aims to find the most frequent set from a collection of instantaneous co-driving sets from multiple timestamps. The mined set must be time-satisfied and scale-satisfied, determined by the two parameters of the shortest co-driving time length $min_t$ and the minimum number of co-driving trucks $min_o$, respectively, that is $|O| \geq min_o \cap |T| \geq min_t$, as shown in Figure 4.

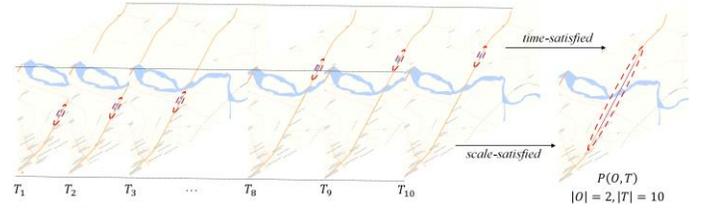

Fig. 4. Schematic of spontaneous platoon pattern mining.

In freight operation, the following distance between trucks could exceed 1 km due to speed difference caused by weigh-in-motion detection and roadside inspection, leading to the interruption of a spontaneous platoon. In this study, $min_t$ is set as 2 timesteps (i.e., 30 seconds), and $min_o$ is set as 2 trucks. To avoid a large number of redundant mining results and eliminate invalid subsets with supersets, the concept of time closure and size closure is introduced. Therefore, the valid spontaneous platoon pattern is defined as those sets that satisfy both time and size closures.

Time Closure: For any spontaneous platoon pattern $P(O, T)$, if no such pattern $P(O', T')$ satisfies $O = O' \cap T \subset T'$, then $P(O, T)$ is called time-closed.

TABLE I
CALCULATED PARAMETERS TO FORM UP A TRUCK PLATOON IN FIGURE 2

|  | $o_1$ | $o_2$ | $o_3$ | $o_4$ | $o_5$ | $o_6$ | $o_7$ | $o_8$ | $o_9$ | $o_{10}$ |
|---|---|---|---|---|---|---|---|---|---|---|
| $y.R$ | Inf | 1.01 | 0.849 | 0.141 | 0.071 | 0.057 | 0.057 | 0.905 | 1.01 | Inf |
| $\theta_{y.R}$ | - | 163 | 142 | 133 | 174 | 178 | 120 | 131 | - | - |
| $\Lambda_{y.R}$ | - | 0.08 | 0.27 | −0.31 | −0.03 | −0.01 | −0.42 | 0.38 | - | - |

Size Closure: For any spontaneous platoon pattern $P(O,T)$, if no such pattern $P(O',T')$ satisfies $O \subset O' \cap T = T'$, then $P(O,T)$ is called size-closed.

For a given set of trucks $O$, it is possible to uniquely determine the maximum co-driving time set $T$, but not vice versa. Therefore, the mining task should apply the approach of depth-first search (DFS) for traversal search to extract the maximum time set $T_{max}(O)$ to ensure that the given trucks are in the same instantaneous co-driving set. The pruning and selection processes are mainly based on the maximum time set $T_{max}(O)$ and can be divided into the following four categories:

Logical Pruning Rule: According to the search order of the depth-first spanning tree, if the difference between the truck index $o_{i_j}$ of the leaf node directly connected to the root and the total number of trucks is less than $min_o$, then the leaf node should be pruned.

A Priori-like Pruning Rule: If the truck set O is not time-satisfied, which means that $T_{max}(O) < min_t$, then the maximum time set corresponding to its all supersets $O'$ is insufficient, that is, $T_{max}(O') < min_t$. None of the child nodes are time-satisfied; therefore, the leaf node where the current truck set $O$ is located should be pruned.

Subset Pruning Rule: According to the searching order of the depth-first spanning tree, if the already found spontaneous platoon pattern $P(O,T)$ and the current pattern $P(O',T')$ satisfy $O \subset O' \cap T' = T$, then the truck set $O'$ and its superset are not time-closed and thus cannot be a valid spontaneous platoon pattern, and the leaf node where the current vehicle set $O'$ is located should be pruned.

Marginal Remove Rule: According to the search order of the depth-first spanning tree, each time the subtree under the root node completes the depth search, nodes that pass through the three pruning process should perform the following marginal remove rule. For the spontaneous platoon pattern $P(O,T)$ in truck set $O = \{o_{i_1}, o_{i_2},\ldots,o_{i_j}\}(i_1 < i_2 < \cdots < i_j)$, if any truck $o_{i_k}(k > j)$ is added and its corresponding pattern $P(O',T')$ satisfies $T' = T$, then the current spontaneous platoon pattern $P(O,T)$ should be removed because it does not satisfy size closure.

*D. Fuel Consumption Estimation*

One of the primary objectives in this study is to evaluate the energy savings incurred from truck platooning. Thus, we apply the longitudinal vehicle model proposed by Liang et al. [22] based on Newton's second law of motion to calculate traction force and then calculate the instantaneous fuel consumption with the fuel consumption model [39], [40], [41]. Truck trajectory data include latitude, longitude, and altitude information, and can be used to compute instantaneous vehicle speed and acceleration rate, which are leading factors that influence truck fuel consumption.

$$ma_t = F_v(t) - F_{airdrag}(v_t) - F_{roll}(\alpha_t) - F_{gravity}(\alpha_t)$$
$$= F_v(t) - \frac{1}{2}\rho A c_d v_t^2 \Phi - mgc_r cos(\alpha_t) - mgsin(\alpha_t)$$
(5)

At any given timestamp $t$, the acceleration rate $a_t$ of a truck with total weight $m$ is mainly determined by engine and braking force $F_v(t)$, roll resistance force $F_{roll}(\cdot)$, gravitational force $F_{gravity}(\cdot)$, and air drag force $F_{airdrag}(\cdot)$. $F_{roll}(\cdot)$ and $F_{gravity}(\cdot)$ are affected by roll resistance coefficient $c_r$, truck weight $m$, gravitational coefficient $g$ and road slope $\alpha_t$ determined by the truck's instantaneous location $z_t$, while $F_{airdrag}(\cdot)$ is influenced by air density $\rho$, the truck's front area A, air drag coefficient $c_d$, instantaneous speed $v_t$, and reduction coefficient $\Phi$. The reduction coefficient $\Phi$ is only less than 1 when the truck is in platoon; otherwise, it is set as 1. The relationship between the instantaneous acceleration and force can be expressed as Equation 5.

Assuming the heat gained from burning the crude oil will be proportionately converted into energy required to propel a truck, the instantaneous fuel consumption $f_c$ (ml/s) can be expressed as Equation 6. $\overline{\eta_{eng}}$ is the mean engine combustion efficiency, $\rho_d$ is the energy conversion constant determined by crude oil, and $\Psi$ is the conversion factor from liquid to gas. $\delta$ is an indicator to prevent oil consumption when the truck is braking, that is, $\delta$ equals 1 when $F_v(t) \geq 0$, and 0 otherwise.

$$f_c = \frac{\delta}{\Psi\overline{\eta_{eng}}\rho_d}F_v(t)v_t \tag{6}$$

Assuming that the acceleration rate of the truck is constant during the 15-second interval $\Delta t$, the total fuel consumption of the truck during each time interval can be expressed as Equation 7.

$$f_c = \int_t^{t+\Delta t}\frac{\delta}{\Psi\overline{\eta_{eng}}\rho_d}F_v(t)v_t dt \tag{7}$$

The driving profile and the changes in speed and altitude of each truck can be acquired from trajectory data. Instantaneous acceleration rate $a_t$ at the moment $t$ based on two consecutive GPS points can be computed through Equation 8.

$$a_t = arctan\left(\frac{v_{t+\Delta t}-v_t}{\Delta t}\right) \tag{8}$$

Instantaneous road slope $\alpha_t$ at the moment $t$ based on two consecutive GPS points can be computed through Equation 9.

$$\alpha_t = arctan\left(\frac{h_{t+\Delta t}-h_t}{d(z_{t+\Delta t}-z_t)}\right) \tag{9}$$

where $h_t$ represents the instantaneous altitude at moment $t$ for each truck, and $d(z_{t+\Delta t} - z_t)$ represents the network distance between two locating points.

Table 2 lists the calibrated parameters for the truck fuel consumption estimation model. Each truck's driving profile (e.g., speed, acceleration rate, and road slope) is computed from the actual trajectory data. The instantaneous axle load data for Liaoning Province within the same year was used to estimate the average total truck weight. The remaining parameters are acquired from the existing literature with identical or similar settings in our study.

TABLE II
PARAMETERS FOR THE FUEL CONSUMPTION MODEL

| Parameter | Unit | Value | Description and Source |
|---|---|---|---|
| $\rho$ | kg/m³ | 1.29 | Liang et al. 2014 |
| $A$ | m² | 10.26 | Liang et al. 2014 |
| $c_d$ | - | 0.007 | Liang et al. 2014 |
| $v_t$ | m/s | - | Derived from trajectory data |
| $\Phi$ | - | 1 | When driving alone; |
|  | - | 0.92 | Leading truck; Lu et al. 2011 |
|  | - | 0.72 | Following truck; Janssen et al. 2015 |
| $m$ | kg | 26,800 | Axle load data in Liaoning Province |
| $g$ | m/s² | 9.8 | - |
| $c_r$ | - | 0.6 | Liang et al. 2014 |
| $\alpha_t$ | - | - | Computed from trajectory data |
| $\delta$ | - | 1 | When $F_v(t) \geq 0$ |
|  | - | 0 | When $F_v(t) < 0$ |
| $\Psi$ | g/ml | 0.737 | Franceschetti et al. 2013 |
| $\overline{\eta_{eng}}$ | - | 0.4 | Industry average level |
| $\rho_d$ | J/g | 44,000 | Franceschetti et al. 2013 |
| $a_t$ | - | - | Compute from trajectory data |

## IV. DATA DESCRIPTION AND PLATOONING PERFORMANCE MEASURES

### A. Data Description

In China, heavy-duty vehicles (HDVs) must be monitored in real time through the fleet management systems governed by transportation authorities. We obtain massive trajectory data from HDVs registered in Liaoning Province on April 10, 2018. The dataset contains 26,405 trucks with 28 million positioning data, of which 20,358 trucks are involved in freight activity. Approximately 77.34% trucks routinely deliver goods within Liaoning Province, while the remaining trucks perform province-to-province freight activities.

GPS devices are equipped on each truck, and they transmit each truck's position information such as truck ID, timestamp, longitude, latitude, altitude, speed, and direction. However, due to signal loss incurred by overhead obstruction and communication failure, the reading frequency is not stable, ranging from 10 s to 60 s in most cases. A certain number of trucks suffer from the no location reporting issue for a short time period but can be properly recovered by the proposed map matching algorithms in Section 3.1. The digital map data come from OSM in December 2019, which include expressway and trunk road network in China. As mentioned in Section 3.1, most trunk road segments are represented in single lines, but trucks can travel in both directions. Figure 5 presents the studied region with truck GPS data.

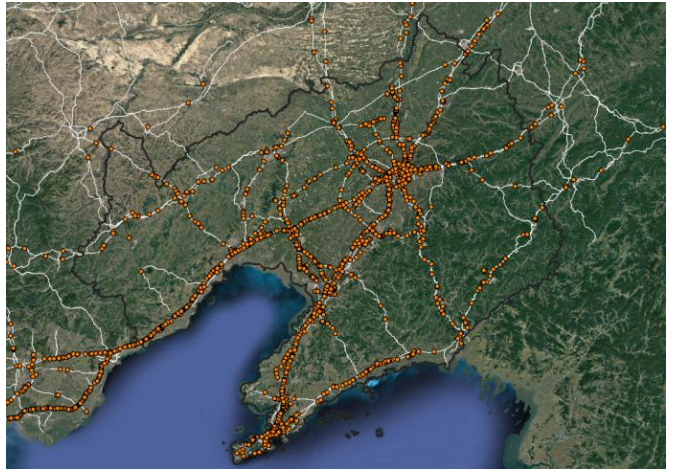

Fig. 5. Truck trajectory distribution in Liaoning province, China.

### B. Platooning Performance Measures

To evaluate the effectiveness of the proposed approaches, we utilize the concept of instantaneous co-driving ratio $ICR_t$ to represent the ratio between the number of mined co-driving trucks and the total available trucks $N_t^{total}$ on the national highway systems at a certain timestamp t. Supposing a total of $S_t$ co-driving sets, and $n_i$ trucks are platooned within each set. Thus, $ICR_t$ can be expressed as

$$ICR_t = \frac{\sum_i^{S_t} n_i}{N^{total}} \quad (10)$$

The space headway between two consecutive coupled trucks is another critical indicator to assess the potential of platooning, which is defined as the average following distance across all instantaneous co-driving sets on a specific road segment at a certain timestamp $t$ (see Equation 11).

$$ICH_t = \frac{\sum_i^{S_t} \sum_j^{n_i-1} h_{ij}}{\sum_i^{S_t} n_i} \quad (11)$$

where $ICH_t$ is the instantaneous co-driving headway, and $h_{ij}$ represents the space headway for *j-th* vehicle within the *i-th* co-driving set.

The average participant trucks in each co-driving sets are an additional indicator to understand the co-driving phenomenon, which is defined as the instantaneous co-driving size $ICS_t$ (see Equation 12).

$$ICS_t = \frac{\sum_i^{S_t} n_i}{S_t} \quad (12)$$

In a macro-level view, if all instantaneous co-driving sets are aggregated, then we can identify trucks' spatial and temporal platooning patterns. Therefore, we propose two indicators to measure the platooning effectiveness in a large freight network. For each individual truck, the percentage of platooning time and distance can be computed and aggregated as the average platooning time and distance ratios for all trucks as defined in Equations 13 and 14.

$$PDR = \frac{\Sigma_j^K PD_j}{\Sigma_j^K D_j} \qquad (13)$$

$$PTR = \frac{\Sigma_j^K PT_j}{\Sigma_j^K T_j} \qquad (14)$$

where $PDR$ and $PTR$ is the average platooning distance and time ratios, respectively; $PD_j$ and $PT_j$ represent the platooned travel distance and time for truck $j$, respectively; $D_j$ and $T_j$ represent the total travel distance and time for truck $j$, respectively; and $K$ indicates the total number of available trucks during the entire day.

## V. RESULT ANALYSIS

### A. Instantaneous Co-driving Set Detection Result

We first apply the map matching algorithm in Section 3.1 to interpolate all trajectory data on the freight network in Liaoning Province and then identify instantaneous co-driving sets by using the proposed A-OPTICS algorithm at each timestamp. As presented in Figure 6, instantaneous truck platooning does not emerge occasionally but rather can be witnessed in major national highways and some road segments around key logistics hubs. The instantaneous co-driving sets can be considered prior signals for proactive truck platoon planning strategies, e.g., speed coordination and route adjustment. Truckers may be informed by fleet management systems to accelerate or decelerate for platoon formation in their next actions.

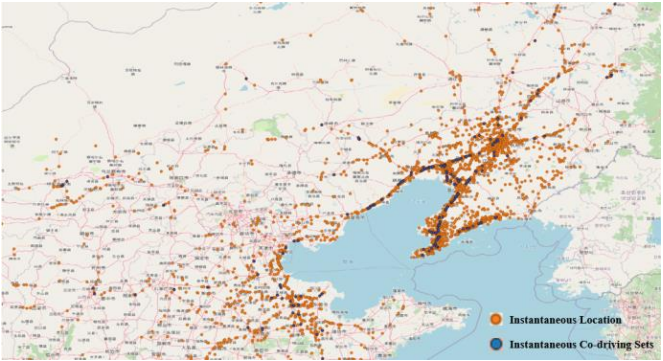

Fig. 6. Spatial distribution of instantaneous co-driving set.

We further compute ICR and ICH by using the trajectory data on April 10, 2018. Figure 7 demonstrates the temporal distribution of an instantaneous co-driving set. Similar to urban traffic congestion patterns, the total available trucks with valid records present a double-peak phenomenon, as presented in Figure 7(a). In contrast, the proportion of trucks on the national highways, including national trunk roads and expressways, remains stable around 50% throughout the entire day, which means that most trucks prefer to travel on highways rather than urban arterial road networks to save time. We divide all trucks on national highways into two categories that exhibit opposite patterns from each other: trucks on trunk roads and trucks on expressways. To be specific, trucks on trunk roads likely deliver goods at night, where more trucks on expressways can be observed during daytime. With the improvement of lamination condition, the ICR values of national trunk roads and expressway present a similar trend of more platooned trucks that can be seen during daytime than during nighttime, especially on trunk roads, as presented in Figures 7(c) and 7(d). ICR and ICS values rise rapidly before morning peak hours (i.e., 5 a.m. to 7 a.m.), whether on national trunk roads or expressways. One noticeable discrepancy between trucks on trunk roads and those on expressways is the ICR value during nighttime. The ICR value of trunk roads become much higher than that of expressways, indicating that trucks are more likely to travel together sporadically on trunk roads. The possible explanation is that there are fewer overloading inspection stations on trunk roads than on expressways and the inspection process is less strict during nighttime than daytime, thus leading to more trucks traveling on trunk roads at night. The size of most truck platoons (i.e., ICS value) during the entire day is 2. The ICS value of trunk roads is slightly higher than that of expressways, implying that more trucks coupled with each other during each trip on trunk roads.

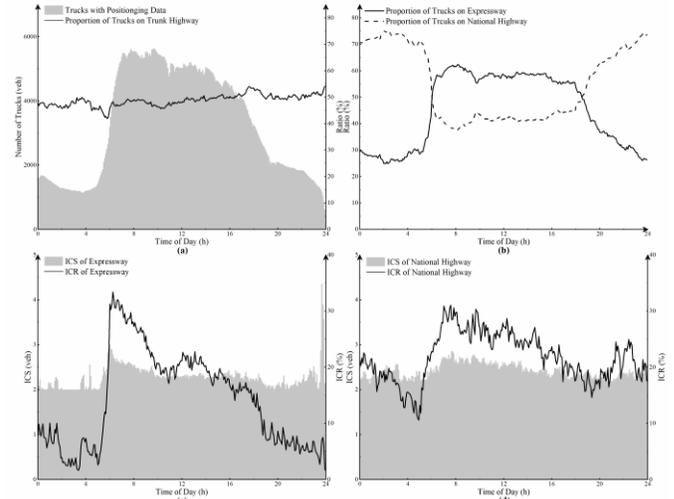

Fig. 7. Temporal distribution of instantaneous co-driving set: (a) proportion of trucks on national highways; (b) proportions of trucks on trunk roads and expressways; (c) ICR and ICS distribution on expressways; (d) ICR and ICS distribution on trunk roads.

By aggregating the 15-second interval into the 5-minute interval, we can further calculate the average headway between two consecutive platooned trucks (i.e., IDH) in Figure 8. The IDH value during daytime is generally more stable than that at nighttime, remaining around 215 and 200 m for expressways and trunk roads, respectively. We find that during nighttime, trucks traveling on trunk roads are more adjacent with each other than those traveling on expressways. This condition is probably due to the higher speed limit on expressways and the presence of more private vehicles at night. When overtaking between trucks and private vehicles occurs, a likely result is the interruption of truck platoons and the fluctuation of headway. This phenomenon becomes more severe during nighttime due to the poor illumination conditions. In contrast, trucks on trunk roads tend to be more cautious during nighttime due to the lower speed limit and fewer private vehicles, leading to a relative stable headway (i.e., ICH) on trunk roads. Considering the small proportion of trucks driving on the expressways and erratic headway fluctuation, performing platoon coordination among trucks traveling on expressways from 7 p.m. to 5 a.m. is not suggested.

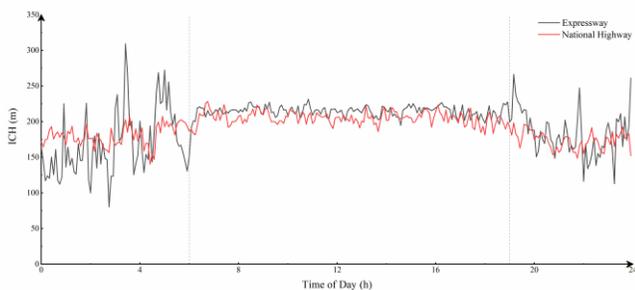

Fig. 8. Temporal distribution of ICH on national trunk roads and expressways.

Under the speed adjustment approach proposed by Liang [22], trucks with relatively small headways can coordinate into a platoon in a relatively short time. From the massive freight trajectory data, trucks' optimal coordination time in the same instantaneous co-driving sets to coordinate into a platoon can be inferred. For national expressways, the best coordination time should be between 5 a.m. to 7 p.m., when the ICR values are relatively high and the ICH values are stable at about 215 m. For national expressways, all timeframes can be considered for truck platooning owing to the relatively stable ICH and considerable ICR values for the entire day.

### B. Spontaneous Platoon Pattern Analysis

We then apply the spontaneous platoon pattern mining algorithm in Section 3.3 to extract the largest truck set with the longest co-driving time period and then analyze the spatiotemporal features and fuel-saving potential by using a freight platoon in Liaoning Province. A typical case of valid spontaneous platoon pattern is shown in Figure 9. Three trucks meet coincidently and participate in the platoon for more than 350 km, with the average instantaneous co-driving headway at around 200 m. The upper-left subgraph presents the zoomed positioning point of three trucks' trajectory data in consecutive timestamps, while the upper-middle subgraph presents the temporal change in the ICR value among the spontaneous platoon pattern during the entire day. Moreover, the spontaneous platoon is interrupted at 9:51 a.m. because one of the trucks switched from the G1 national expressway to the G102 trunk road. However, the three trucks finally meet by chance on Tianjin's expressway and continue to spontaneously travel as a group for a while. This example illustrates that the proposed algorithms can well capture the trucks' accidental meeting behaviors and identify spontaneous platooning patterns in the large-scale freight network.

To further study the feature of mined spontaneous platoon patterns, we depict the platooning time and distance in Figure 10. Nearly 30% trucks are platooned for at least 10 minutes and travel as a group for 10 km.

In early 2013, Liang [19] revealed that when two trucks are traveling on the same route but with a relatively short headway, a platoon can be formed up via speed coordination for fuel savings as long as their overlapping distance is more than 17 times their headway. The above theoretical foundations can be utilized with the mined spontaneous platoon patterns in this study to further estimate fuel savings. From the micro-level perspective, 52.54% of the trucks on national highways (i.e., both expressways and trunk roads) co-drive with other trucks for longer or shorter durations, and the average platooning distance PDR and the average time ratios PTR for these platooned trucks in Liaoning Province are 9.645% and 9.943%, respectively. By jointly considering the average truck headway and overlapping travel distance among all mined platooning patterns, 35.83% patterns can undergo speed coordination for platooning, resulting in a potential 2.767% reduction in total fuel consumption for trucks on national highways on April 10, 2018. The ratio of potential fuel savings is similar to that reported by Muratori et al. [42], where 4.2% of fuel savings can be achieved through truck platooning.

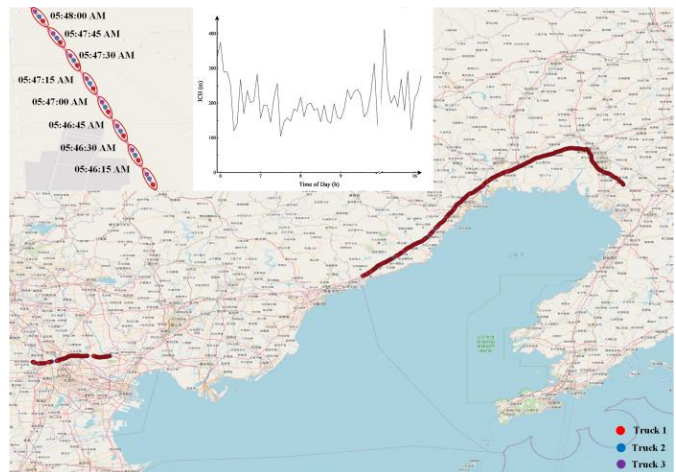

Fig. 9. Typical case of valid spontaneous platoon pattern with three trucks.

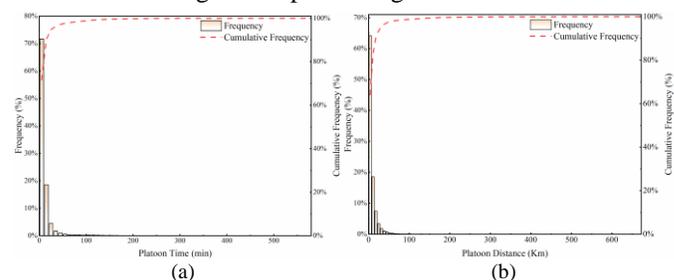

Fig. 10. Characteristics of spontaneous truck platoon patterns: (a) platooning time distribution; (b) platooning distance distribution.

Common sense suggests that trucks with long-distance travels on national highways are more likely to be platooned with higher energy saving potential, and our findings directly support this assumption. We further outline the relationship between truck daily hauling distance and platooning indicators (i.e., PDR and PTR) in Figure 11. In general, a longer daily hauling distance implies a higher opportunity for platoon coordination. However, the PDR and PTR values gradually decrease when the daily hauling distance exceeds 1000 km because trucks with relatively long hauling distances tend to participate in interprovincial deliveries, leading to lower platooning performance measures. This issue can be further tested by analyzing the nationwide trucks' trajectory data.

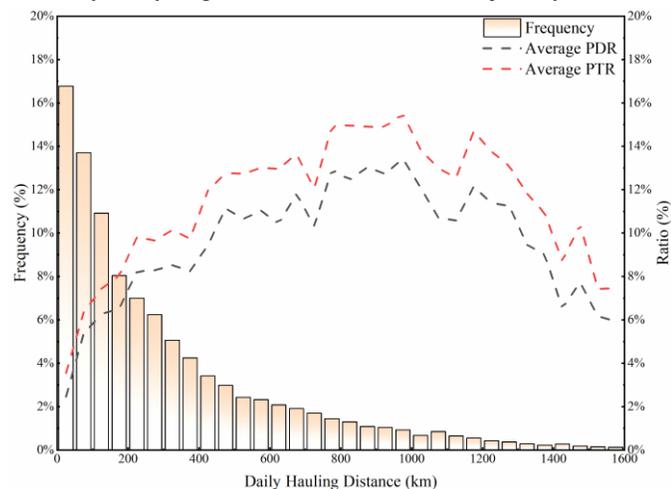

Fig. 11. Spontaneous truck platooning performance with different daily hauling distances.

Spontaneous truck platoon patterns exhibit a certain spatial regularity. Therefore, installing roadside communication units on those segments where platooned trucks routinely travel is more cost efficient than network-wide infrastructure constructions. We summarize and visualize the occurrences of truck platoons in Figure 12; the segments with high platooning potential are marked in red. Most of these segments are adjacent to industrial manufacturers and port logistics. The segments with the highest possibility of platooning are in the trunk road G228 and G305, which are the major corridors of Panjin Port, one of the largest ports in northern China. More spontaneous platooning patterns are observed in the Panjin–Jinzhou section of the G1 expressway and Yingkou–Dalian section of the G15 expressway. The G1 section connects Jinzhou, the largest logistics center of Liaoning Province, with Panjin, the base of petroleum and petrochemical industry. The G15 expressway facilitates the cargo flow between the two major harbor cities of Liaoning Province (i.e., Yinkou and Dalian). As a result of the radiation effect of Yingkou Port, a high proportion of spontaneous platoon patterns occur in the G102 trunk roads, which serve as a juncture of the G1 national expressway. As a result, the site selection of the marshalling station for assembling truck platoons can be prioritized based on the spatial distribution of spontaneous platoon patterns.

## VI. Conclusions

In the past decade, truck platoon planning has shifted from theoretical derivation toward closed road test. The fuel consumption and air pollution incurred by truck platooning can be significantly reduced. To reveal the potential of platooning in large-scale freight networks, a series of data mining approaches is proposed to mine spontaneous platooning patterns from massive truck trajectory data.

A map matching approach with heading detection is developed to identify the relative direction of a truck traveling on a single-line bidirectional road segment. An enhanced OPTICS algorithm is further proposed to recognize the instantaneous co-driving sets at each timestamp. This algorithm overcomes the issue of mistaken clustering by replacing the conventional two-dimensional Euclidean distance with one-dimensional vehicle following distance, thus being able to detect the correct convoy relationship among multiple trucks by considering the change in reachability distances. We further aggregate multiple instantaneous co-driving sets and adopt the frequent itemset mining with pruning rules to find spontaneous platoon patterns that satisfy platoon size and duration constraints. Fuel savings due to truck platooning are then computed by using a well-established energy consumption estimated model.

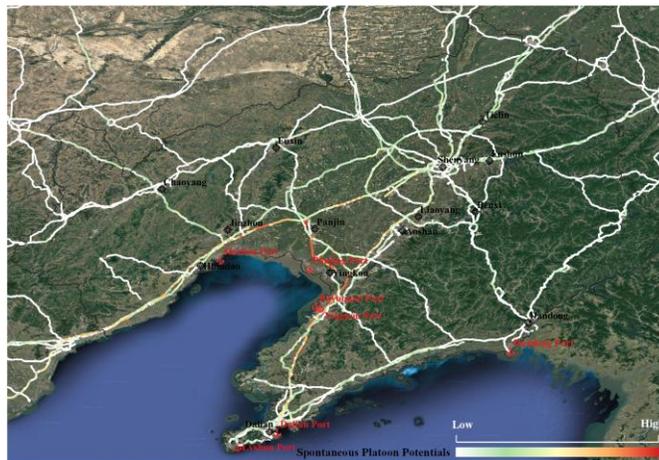

Fig. 12. Road segments with spontaneous truck platooning potential in Liaoning Province.

To validate the effectiveness of the proposed truck platoon pattern mining approaches, we leverage the extensive truck trajectory data from 26,405 trucks with 28.3345 million GPS records registered in Liaoning Province, China, on April 10, 2018. By mapping with OSM digital map data, we can find that at least 52.54% of trucks spontaneously coupled with each other for a while on a given segment, and the average platooning distance and time ratios are 9.645% and 9.943%, respectively. Among the spontaneous platoon patterns, 35.83% of the patterns need proper speed coordination for platooning, and the reduced energy savings could be as high as 2.767%. The majority of trucks prefer either national freeways or trunk roads. Thus, we further calculate the instantaneous co-driving ratio, size, and distance for both road types and find that the trucks are more likely to be platooned during daytime than during nighttime, with an average platooning distance of around 200 m. The co-driving ratio of trunk roads is higher than that of national freeways probably because of the lower speed limits on those trunk roads and fewer private vehicles at night, causing less intervention with truck platoons. We also reveal that the average platooning distance and time ratios are positively related to each truck's daily haul distance when the average haul distance is less than 1000 km but decline when a longer distance travel is observed because trucks may undertake interprovincial deliveries. The road segments with high platooning potential are highlighted in the map. Most spontaneous truck platoons can be found in locations adjacent to ports and logistics hubs.

On the basis of the mined patterns, the following policy

implications can be made to better guide autonomous truck platooning implementation:

(1) Platooning strategies could be time dependent. Speed coordination can be executed from 5 a.m. to 7 p.m. for trucks traveling on national freeways and is available for trucks on trunk roads to be platooned.

(2) The typical space headway for spontaneous truck platoon planning is 200 m. Thus, trucks may be equipped with medium-range communication devices to coordinate with another following or leading truck for platooning.

(3) Almost half of the trucks that are traveling spontaneously in Liaoning Province can be readily platooned through speed coordination without changing their scheduling and routes.

(5) The investment priority should be for roadside communication infrastructure in areas where truck platoons more likely occur. Taking Liaoning Province as an example, the G228 and G305 national freeways linking to Yingkou Port, the Panjin–Jinzhou section in the G1 expressway, and the Yingkou–Dalian section in the G15 expressway should consider installing V2X communication equipment.

Future research effort can be made in the following aspects. Long-distance travels for interprovincial deliveries should be investigated to assess truck platooning potential. The efficiency of the A-OPTICS algorithm for instantaneous co-driving set detection is not satisfactory; the algorithm needs to be parallelized. A future study should focus on optimizing speed coordination and scheduling alternations on the basis of the mined spontaneous truck platooning patterns.

APPENDIX

A. *Pseudocode for the Enhanced OPTICS Algorithm*

1) *Algorithm FindValley*

```
Algorithm    FindValley
Input:  Array Core Distance CD
Output: Array Valley Sets VS
1: function FINDVALLEY(CoreDistance CD)
2:     Candidatelist ← range(0,len(CD))[CD < 1]
3:     stloc, stloc_ ← Candidatelist[0]
4:     Array Valley Sets VS
5:     for i = 1 → len(Candidatelist) do
6:         if Candidatelist[i] - stloc_ = 1 then
7:             stloc_ ← stloc_ + 1
8:         else
9:             edloc ← Candidatelist[i-1]
10:            VS.append(range(stloc-1,edloc+2))
11:            stloc, stloc_ ← Candidatelist[i]
12:        end if
13:    end for
14:    return VS
15: end function
```

ATTACHED Fig. 1. Function FindValley.

2) *Function Adaptive Recognition*

```
Algorithm    Adaptive Recognition
Input:  Array Reachability Distance RD, Array Valley Sets VS, Interget Eps, Interget MinPTs, Integer θ', Float Δ
Output: Array Co-driving Sets CS
1: function ADAPTIVE RECOGNITION(ValleySets VS, ReachabilityDistance RD, MinPTs, Eps, θ', Δ)
2:     Co - drivingSets CS ← [ ]
3:     for i = 1 → len(VS) do
4:         Array Valley ReachabilityDistance VR ← RD[VS[i]]
5:         VR[1],VR[-2] ← 1.01
6:         SetStartPts ← [ ]
7:         TargetSet ← [ ]
8:         TempSet ← [ ]
9:         TempSetEndPT ← len(VR)-1
10:        SetStartPts.append(VS[i][0])
11:        for j = 1 → len(VR) - 1 do
12:            if Θ_{j,k}<θ' then
13:                if Δ_{j,k}<0 then
14:                    if len(TempSet)>MinPTs then
15:                        TargetSet.append(TempSet)
16:                    end if
17:                    TempSet ← [ ]
18:                    if len(SetStartPts)>0 then
19:                        while RD[SetStartPts[-1]]<RD[j] do
20:                            TempSet ← range(SetStartPts[-1],TempSetEndPT)
21:                            if len(TempSet)>MinPTs then
22:                                TargetSet.append(TempSet)
23:                            end if
24:                            SetStartPts.pop()
25:                        end while
26:                        if len(TempSet)>MinPTs then
27:                            TargetSet.append(range(SetStartPts[-1],TempSetEndPT))
28:                        end if
29:                    end if
30:                    if RD[j + 1]<RD[j] then
31:                        SetStartPts.append(j)
32:                    end if
33:                else
34:                    if VR[j + 1]>VR[j] then
35:                        TempSetEndPT ← j + 1
36:                        TempSet ← range(SetStartPts[-1],TempSetEndPT)
37:                        if j + 1 = TempSetEndPT & VR[j + 1]>Eps & VR[j]<Eps then
38:                            TargetSet.append(TempSet)
39:                        end if
40:                    end if
41:                end if
42:            end if
43:        end for
44:        if len(TargetSet) =0 then
45:            CSpart = range(SetStartPts[-1],len(VR) - 1)
46:            CS.append(range(min(CSpart),max(CSpart)+2))
47:        else
48:            while len(SetStartPts)>0 do
49:                TempSet ← range(SetStartPts[-1],len(VR) - 1)
50:                if RD[SetStartPts[-1]]>RD[-1] & len(TempSet)>MinPTs then
51:                    TargetSet.append(TempSet)
52:                end if
53:                SetStartPts.pop()
54:            end while
55:            Droplist ← [VR[1],VR[-1]]
56:            for k = 0 → len(TargetSet) do
57:                CSpart ← TargetSet[k]
58:                if CSpart[0] notin Droplist & CSpart[-1] notin Droplist then
59:                    CS.append(CSpart)
60:                end if
61:            end for
62:        end if
63:    end for
64:    return Co - drivingSets CS
65: end function
```

ATTACHED Fig. 2. Function Adaptive Recognition

3) *Algorithm A-OPTICS*

```
Algorithm    A-OPTICS
Input:  Array Temporary Location TL, Interget Eps, Interget MinPTs, Integer θ', Float Δ
Output: Array Co-driving Sets CS
1: ReachabilityDistance RD,CoreDistance CD ← P-OPTICS(TL, Eps, MinPTs)
2: ValleySets VS ← FindValley(CD)
3: Co - drivingSets CS ← Adaptive Recognition(VS,RD, MinPTs, Eps, θ', Δ)
```

ATTACHED Fig. 3. Algorithm A-OPTICS.

B. *Calculation Approach for Following Distance*

To compute the following distance between two trucks, whether they are in following relationships first needs to be determined. The intuitive idea is to use the difference of positioning and heading data of two trucks in consecutive timestamps for comprehensive judgment. However, due to the inherent error of recorded heading and the variety of road distribution, we cannot find a threshold to judge the following relationship accurately. Fortunately, the matched segment $Segment_{o_i}$ of truck $o_i$ combined with its relative driving direction $dir_{o_i}$ and the headed node $ToNode_{o_i}$ can help in accurately recognizing the following relationship. Here, $r_{o_i}$ is the ratio of the length from the segment's start node to the matched locations $o_i$ to the total segment length.

The most straightforward situation is that two trucks are traveling in the same segment at a given moment, so the relative driving direction will directly help determine whether they are in a following relationship and thus, the following distance can be calculated. When the relative driving directions are the same, the following distance $FD(o_1, o_2)$ can be determined according to the difference in their location $r_{o_1}$ and $r_{o_2}$ combined with the segment length, as shown in Table 1. However, when the relative driving directions are opposite, they cannot follow at this moment, and the following distance can be set to infinity.

However, segments that make up the roads in the digital map have major differences in length. The road with $\varepsilon$-length can be represented by any single or dual lines or in combination with

ATTACHED TABLE I
FOLLOWING DISTANCE FOR TRUCKS ON THE SAME SEGMENT

| | Scenario 1 | Scenario 2 |
|---|---|---|
| | 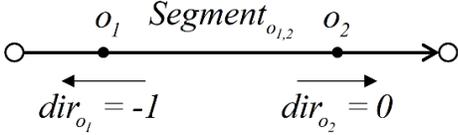 | 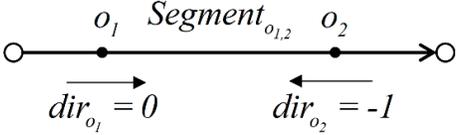 |
| Features $FD(o_1, o_2)$ | $dir_{o_1} \neq dir_{o_2}$<br>$+\infty$ | $dir_{o_1} \neq dir_{o_2}$<br>$+\infty$ |
| | Scenario 3 | Scenario 4 |
| | 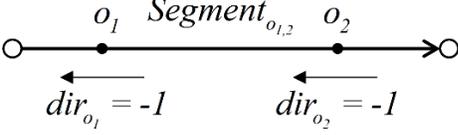 | 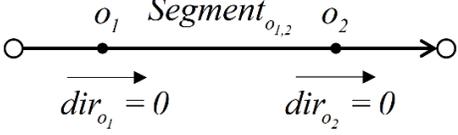 |
| Features $FD(o_1, o_2)$ | $dir_{o_1} = dir_{o_2}$<br>$Seglen \cdot |r_{o_1} - r_{o_2}|$ | $dir_{o_1} = dir_{o_2}$<br>$Seglen \cdot |r_{o_1} - r_{o_2}|$ |

ATTACHED TABLE II
FOLLOWING DISTANCE FOR TRUCKS ON DIFFERENT SINGLE SEGMENTS

| | Scenario 1 | Scenario 2 |
|---|---|---|
| | 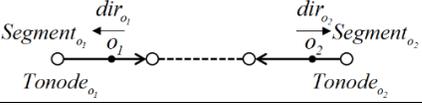 | 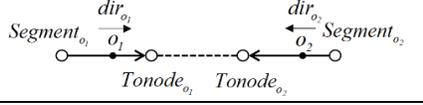 |
| $ETE_{path}(o_1, o_2)$ | 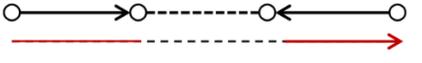 | 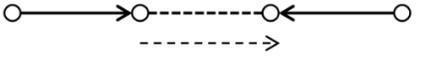 |
| $ETE_{path}(o_2, o_1)$ | 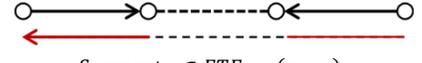 | 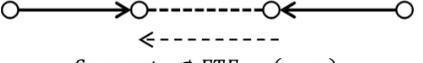 |
| Features | $Segment_{o_1} \in ETE_{path}(o_1, o_2)$,<br>$Segment_{o_1} \in ETE_{path}(o_2, o_1)$,<br>$Segment_{o_2} \in ETE_{path}(o_1, o_2)$,<br>$Segment_{o_2} \in ETE_{path}(o_2, o_1)$ | $Segment_{o_1} \notin ETE_{path}(o_1, o_2)$,<br>$Segment_{o_1} \notin ETE_{path}(o_2, o_1)$,<br>$Segment_{o_2} \notin ETE_{path}(o_1, o_2)$,<br>$Segment_{o_2} \notin ETE_{path}(o_2, o_1)$ |
| $FD(o_1, o_2)$ | $+\infty$ | $+\infty$ |
| | Scenario 3 | Scenario 4 |
| | 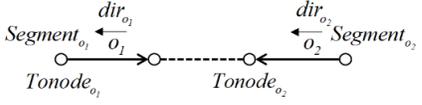 | 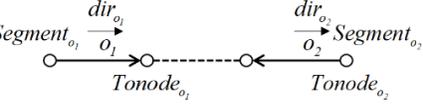 |
| $ETE_{path}(o_1, o_2)$ | 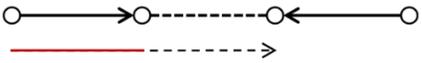 | 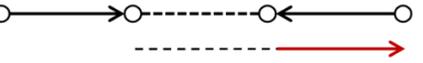 |
| $ETE_{path}(o_2, o_1)$ | 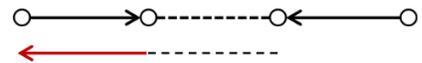 | 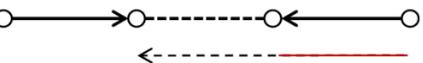 |
| Features | $Segment_{o_1} \in ETE_{path}(o_1, o_2)$,<br>$Segment_{o_1} \in ETE_{path}(o_2, o_1)$ | $Segment_{o_2} \in ETE_{path}(o_1, o_2)$,<br>$Segment_{o_2} \in ETE_{path}(o_2, o_1)$ |
| $FD(o_1, o_2)$ | $ETE_{dis}(o_2, o_1) + Edgelen_{o_2} \times \theta_{o_2} - Edgelen_{o_1} \times \theta_{o_1}$ | $ETE_{dis}(o_1, o_2) + Edgelen_{o_1} \times \theta_{o_1} - Edgelen_{o_2} \times \theta_{o_2}$ |

both segments, as shown in Attached Tables 2 to 4. As a result, not all the considered co-driving trucks will be located in the same segment, which poses a challenge to recognizing the following relationship and determining the following distance. To reduce the calculation time, only the two trucks whose geographical distance is lower than $\varepsilon$ are judged.

Therefore, when two trucks are not located on the same segment, we introduced the end-to-end route $ETE_{path}(o_1, o_2)$ and end-to-end distance $ETE_{dis}(o_1, o_2)$ as defined below.

**End-to-end Route** $ETE_{path}(o_1, o_2)$: The end-to-end route $ETE_{path}(o_1, o_2)$ for truck $o_1$ to truck $o_2$ is defined as the shortest network route from $ToNode_{o_1}$ to $ToNode_{o_2}$ in a national road network consisting of national freeways and trunk roads.

**End-to-end Distance** $ETE_{dis}(o_1, o_2)$: The end-to-end distance $ETE_{dis}(o_1, o_2)$ for truck $o_1$ to truck $o_2$ is defined as the length of the end-to-end route $ETE_{path}(o_1, o_2)$.

Considering that U-turn is not allowed for trucks on the national road, the catch-up distance $CD(o_1, o_2)$ of truck $o_1$ to truck $o_2$ can be calculated as $CD(o_1, o_2) = Dis(o_1, ToNode_{o_1}) + ETE_{dis}(o_1, o_2) + Dis(o_2, ToNode_{o_2})$, and the following distance $FD(o_1, o_2)$ between two trucks can be calculated as $FD(o_1, o_2) = min(CD(o_1, o_2), CD(o_2, o_1))$. The following distance has symmetry, that is, $FD(o_1, o_2) =$

ATTACHED TABLE III
FOLLOWING DISTANCE FOR TRUCKS ON DIFFERENT DUAL SEGMENTS

| | Scenario 1 | Scenario 2 |
|---|---|---|
| $ETE_{path}(o_1, o_2)$ | | |
| $ETE_{path}(o_2, o_1)$ | | |
| Features | $ETE_{dis}(o_1, o_2) \gg \varepsilon, ETE_{dis}(o_2, o_1) \gg \varepsilon$ | $ETE_{dis}(o_2, o_1) \gg \varepsilon$ |
| $FD(o_1, o_2)$ | $+\infty$ | $ETE_{dis}(o_1, o_2) + Edgelen_{o_1} \times \theta_{o_1} - Edgelen_{o_2} \times \theta_{o_2}$ |

ATTACHED TABLE IV
FOLLOWING DISTANCE FOR TRUCKS ON DIFFERENT SEGMENTS

| | Scenario 1 | Scenario 2 |
|---|---|---|
| $ETE_{path}(o_1, o_2)$ | | |
| $ETE_{path}(o_2, o_1)$ | | |
| Features | $ETE_{dis}(o_2, o_1) \gg \varepsilon$, $Segment_{o_2} \in ETE_{path}(o_1, o_2)$, $Segment_{o_1} \notin ETE_{path}(o_1, o_2)$ | $ETE_{dis}(o_2, o_1) \gg \varepsilon$, $Segment_{o_2} \in ETE_{path}(o_1, o_2)$, $Segment_{o_1} \in ETE_{path}(o_1, o_2)$ |
| $FD(o_1, o_2)$ | $ETE_{dis}(o_1, o_2) + Edgelen_{o_1} \times \theta_{o_1} - Edgelen_{o_2} \times \theta_{o_2}$ | $+\infty$ |
| | Scenario 3 | Scenario 4 |
| $ETE_{path}(o_1, o_2)$ | | |
| $ETE_{path}(o_2, o_1)$ | | |
| Features | $ETE_{dis}(o_1, o_2) \gg \varepsilon$, $Segment_{o_2} \notin ETE_{path}(o_2, o_1)$, $Segment_{o_1} \notin ETE_{path}(o_2, o_1)$ | $ETE_{dis}(o_1, o_2) \gg \varepsilon$, $Segment_{o_2} \notin ETE_{path}(o_2, o_1)$, $Segment_{o_1} \in ETE_{path}(o_2, o_1)$ |
| $FD(o_1, o_2)$ | $+\infty$ | $ETE_{dis}(o_2, o_1) + Edgelen_{o_2} \times \theta_{o_2} - Edgelen_{o_1} \times \theta_{o_1}$ |

$FD(o_2, o_1)$. In the Attached Tables 2 to 4, the $\theta_{o_i}$ represents the ratio of the remaining distance, and the truck $o_i$ travels in the current driving direction to its heading node $ToNode_{o_1}$ to the length of the current segment, which can be obtained from Attached Equation 1.

$$\theta_{o_i} = \begin{cases} 1 - r_{o_i} & dir_{o_i} = 0 \\ r_{o_i} & dir_{o_i} = -1 \end{cases} \quad (1)$$

Attached Table 2 shows all the possibilities of determining whether two trucks are co-driving on the trunk road presented by single bidirectional segments and close to each other. The dotted line refers to the segments that connect $Segment_{o_1}$ and $Segment_{o_2}$ where the trucks are located. Obviously, the catch-up distance $CD(o_1, o_2)$ for two trucks driving face to face or back to back may tend to be smaller than $\varepsilon$ caused by the two-way movement of bidirectional segments, thus resulting in the misidentification of the following relationship, such as scenarios 1 and 2 in Attached Table 2. Fortunately, the end-to-end route $ETE_{path}(o_1, o_2)$ may effectively assist in eliminating the misidentification results. As shown in Attached Table 2, when the two trucks are in a following relationship, the end-to-end route $ETE_{path}(o_1, o_2)$ and $ETE_{path}(o_2, o_1)$ will separately contain and only contain the segment where the leading truck is located. Therefore, the following distance could be calculated when two trucks satisfy this feature. Otherwise, it should be set to infinity.

Attached Table 3 shows all the possibilities of determining whether two trucks are co-driving on the national freeway presented by dual unidirectional segments and whether they are close to each other. When two trucks are not in a following relationship, the end-to-end distance $ETE_{dis}(o_1, o_2)$ and $ETE_{dis}(o_2, o_1)$ would be significantly higher than $\varepsilon$. In contrast, only one of the end-to-end distances of two trucks would be significantly higher than $\varepsilon$ when they are co-driving together. Therefore, the following distance of paired trucks that satisfy the above feature should be calculated, while others' following distance should be directly set to infinity. The conclusion is also correct when the positions of $o_1$ and $o_2$ are replaced.

In general, the conversion of single bidirectional segments presented road and the dual unidirectional segments presented road will occur at the junction of the national freeway and the trunk road. Therefore, an essential step is to discuss the four possibilities of determining whether two trucks are co-driving when they are separately located on bidirectional and unidirectional segments. The research background is the national road network; thus, when the paired trucks are co-driving together, the end-to-end distance's and end-to-end route's features are unique. The leading truck's end-to-end distance would be significantly higher than $\varepsilon$, while the end-to-end route of the following truck would contain the segment where the leading truck is located but not the segment where it is located. The above features are shown in scenarios 1 and 4 in Attached Table 4. The paired trucks that satisfy these features are co-driving together, and their following distance needs to be calculated.

## References

[1] OECD, ITF Transport Outlook 2013: Funding Transport. Paris: OECD Publishing, 2013.
[2] B. M. Schroten A, Warringa G, "Marginal abatement cost curves for heavy duty vehicles," Delft, 2012.
[3] C. Bergenhem, H. Pettersson, E. Coelingh, C. Englund, S. Shladover, and S. Tsugawa, "Overview of platooning systems," 2012.
[4] B. Patten, J., McAucliffe, B., Mayda, W. & Tanguay, "Review of Aerodynamic Drag Reduction Devices for Heavy Trucks and Buses," Ottawa, 2012.
[5] A. Davila, E. Aramburu, and A. Freixas, "Making the best out of aerodynamics: Platoons," in SAE Technical Papers, 2013, vol. 2, doi: 10.4271/2013-01-0767.
[6] B. Van Arem, C. J. G. Van Driel, and R. Visser, "The impact of cooperative adaptive cruise control on traffic-flow characteristics," IEEE Trans. Intell. Transp. Syst., vol. 7, no. 4, pp. 429–436, Dec. 2006, doi: 10.1109/TITS.2006.884615.
[7] J. Lioris, R. Pedarsani, F. Y. Tascikaraoglu, and P. Varaiya, "Platoons of connected vehicles can double throughput in urban roads," Transp. Res. Part C Emerg. Technol., vol. 77, pp. 292–305, Apr. 2017, doi: 10.1016/j.trc.2017.01.023.
[8] F. Zhu and S. V. Ukkusuri, "Modeling the Proactive Driving Behavior of Connected Vehicles: A Cell-Based Simulation Approach," Comput. Civ. Infrastruct. Eng., vol. 33, no. 4, pp. 262–281, Apr. 2018, doi: 10.1111/mice.12289.
[9] D. Zhang, Z. Xu, D. Srinivasan, and L. Yu. "Leader–Follower Consensus of Multiagent Systems With Energy Constraints: A Markovian System Approach", IEEE Transactions on Cybernetics, vol. 47, no. 7. pp. 1727–1736, July. 2017, doi: 10.1109/TSMC.2017.2677471.
[10] Z. Zhang, L. Zhang, F. Hao, and L. Wang. "Leader-Following Consensus for Linear and Lipschitz Nonlinear Multiagent Systems With Quantized Communication", IEEE Transactions on Cybernetics, vol. 47, no. 8. pp. 1970–1982, Aug. 2017, doi: 10.1109/TCYB.2016.2580163.
[11] G. Wen, C.L.P. Chen, Y.-J.. Liu, and Z. Liu. "Neural Network-Based Adaptive Leader-Following Consensus Control for a Class of Nonlinear Multiagent State-Delay Systems", IEEE Transactions on Cybernetics, vol. 47, no. 8. pp. 2151–2160, Aug. 2017, doi: 10.1109/TCYB.2016.2608499.
[12] X. Tan, J. Cao, amd X. Li. "Consensus of Leader-Following Multiagent Systems: A Distributed Event-Triggered Impulsive Control Strategy", IEEE Transactions on Cybernetics, vol. 49, no. 3. pp. 792-801, March. 2019, 10.1109/TCYB.2017.2786474.
[13] X. You, C. Hua, and X. Guan. "Self-Triggered Leader-Following Consensus for High-Order Nonlinear Multiagent Systems via Dynamic Output Feedback Control", IEEE Transactions on Cybernetics, vol. 49, no. 6. pp. 2002-2010, June. 2019, 10.1109/TCYB.2018.2813423.
[14] D. Yue and Z. Meng. "Cooperative Set Aggregation of Second-Order Multiagent Systems: Approximate Projection and Prescribed Performance", IEEE Transactions on Cybernetics., vol. 50, no. 3. pp. 957-970, March. 2020, doi: 10.1109/TCYB.2018.2875131.
[15] Z. Liu, D. Li, L., Wang, and D. Dong. "Synchronization of a Group of Mobile Agents With Variable Speeds Over Proximity Nets", IEEE Transactions on Cybernetics, vol. 46, no. 7. pp. 1579-1590, July. 2016, doi: 10.1109/TCYB.2015.2451695.
[16] A. Petrillo, A. Pescapé, and S. Santini. "A Secure Adaptive Control for Cooperative Driving of Autonomous Connected Vehicles in the Presence of Heterogeneous Communication Delays and Cyberattacks", IEEE Transactions on Cybernetics, in press, Janurary. 2020, doi: 10.1109/TCYB.2019.2962601.
[17] A. K. Bhoopalam, N. Agatz, and R. Zuidwijk, "Planning of truck platoons: A literature review and directions for future research," Transportation Research Part B: Methodological, vol. 107. Elsevier Ltd, pp. 212–228, Jan. 01, 2018, doi: 10.1016/j.trb.2017.10.016.
[18] V. Reis, R. Pereira, and P. Kanwat, "Assessing the potential of truck platooning in short distances: The case study of Portugal," in Urban Freight Transportation Systems, Elsevier, 2019, pp. 203–222.
[19] K. Y. Liang, J. Martensson, and K. H. Johansson, "When is it fuel eficient for a heavy duty vehicle to catch up with a platoon?," in IFAC Proceedings Volumes (IFAC-PapersOnline), 2013, vol. 7, no. PART 1, pp. 738–743, doi: 10.3182/20130904-4-JP-2042.00071.
[20] S. van de Hoef, "Fuel-Efficient Centralized Coordination of Truck Platooning." Jan. 01, 2016, Accessed: Sep. 23, 2020. [Online]. Available: https://www.mysciencework.com/publication/show/fuel-efficient-centralized-coordination-truck-platooning-4855776d.
[21] W. Zhang, E. Jenelius, and X. Ma, "Freight transport platoon coordination and departure time scheduling under travel time uncertainty," Transp. Res. Part E Logist. Transp. Rev., vol. 98, pp. 1–23, Feb. 2017, doi: 10.1016/j.tre.2016.11.008.
[22] K. Y. Liang, J. Mårtensson, and K. H. Johansson, "Heavy-Duty Vehicle Platoon Formation for Fuel Efficiency," IEEE Trans. Intell. Transp. Syst., vol. 17, no. 4, pp. 1051–1061, Apr. 2016, doi: 10.1109/TITS.2015.2492243.
[23] F. Farokhi and K. H. Johansson, "Investigating the interaction between traffic flow and vehicle platooning using a congestion game," in IFAC Proceedings Volumes (IFAC-PapersOnline), 2014, vol. 19, pp. 4170–4177, doi: 10.3182/20140824-6-za-1003.00847.

[24] P. Meisen, T. Seidl, and K. Henning, "A data-mining technique for the planning and organization of truck platoons," in International Conference on Heavy Vehicles HVParis 2008, Hoboken, NJ, USA: John Wiley & Sons, Inc, 2013, pp. 389–402.
[25] J. Larson, C. Kammer, K. Y. Liang, and K. H. Johansson, "Coordinated route optimization for heavy-duty vehicle platoons," in IEEE Conference on Intelligent Transportation Systems, Proceedings, ITSC, 2013, pp. 1196–1202, doi: 10.1109/ITSC.2013.6728395.
[26] E. Larsson, G. Sennton, and J. Larson, "The vehicle platooning problem: Computational complexity and heuristics," Transp. Res. Part C Emerg. Technol., vol. 60, pp. 258–277, Nov. 2015, doi: 10.1016/j.trc.2015.08.019.
[27] J. Larson, T. Munson, and V. Sokolov, "Coordinated Platoon Routing in a Metropolitan Network," in 2016 Proceedings of the Seventh SIAM Workshop on Combinatorial Scientific Computing, Philadelphia, PA: Society for Industrial and Applied Mathematics, 2016, pp. 73–82.
[28] A. Nourmohammadzadeh and S. Hartmann, "The fuel-efficient platooning of heavy duty vehicles by mathematical programming and genetic algorithm," in Lecture Notes in Computer Science (including subseries Lecture Notes in Artificial Intelligence and Lecture Notes in Bioinformatics), Dec. 2016, vol. 10071 LNCS, pp. 46–57, doi: 10.1007/978-3-319-49001-4_4.
[29] V. Pillac, M. Gendreau, C. Guéret, and A. L. Medaglia, "A review of dynamic vehicle routing problems," Eur. J. Oper. Res., vol. 225, no. 1, pp. 1–11, Feb. 2013, doi: 10.1016/j.ejor.2012.08.015.
[30] A. Adler, D. Miculescu, and S. Karaman, "Optimal Policies for Platooning and Ride Sharing in Autonomy-Enabled Transportation," Springer, Cham, 2020, pp. 848–863.
[31] H. Jeung, M. L. Yiu, X. Zhou, C. S. Jensen, and H. T. Shen, "Discovery of convoys in trajectory databases," Proc. VLDB Endow., vol. 1, no. 1, pp. 1068–1080, Aug. 2008, doi: 10.14778/1453856.1453971.
[32] K. Y. Liang, J. Martensson, and K. H. Johansson, "Fuel-saving potentials of platooning evaluated through sparse heavy-duty vehicle position data," in IEEE Intelligent Vehicles Symposium, Proceedings, 2014, pp. 1061–1068, doi: 10.1109/IVS.2014.6856540.
[33] J. Larson, K. Y. Liang, and K. H. Johansson, "A distributed framework for coordinated heavy-duty vehicle platooning," IEEE Trans. Intell. Transp. Syst., vol. 16, no. 1, pp. 419–429, Feb. 2015, doi: 10.1109/TITS.2014.2320133.
[34] T. T. Shein, S. Puntheeranurak, and M. Imamura, "Discovery of evolving companion from trajectory data streams," Knowl. Inf. Syst., vol. 62, no. 9, pp. 3509–3533, Sep. 2020, doi: 10.1007/s10115-020-01471-2.
[35] M. Andersson, J. Gudmundsson, P. Laube, and T. Wolle, "Reporting leaders and followers among trajectories of moving point objects," Geoinformatica, vol. 12, no. 4, pp. 497–528, Oct. 2008, doi: 10.1007/s10707-007-0037-9.
[36] P. Newson and J. Krumm, "Hidden Markov map matching through noise and sparseness," in GIS: Proceedings of the ACM International Symposium on Advances in Geographic Information Systems, 2009, pp. 336–343, doi: 10.1145/1653771.1653818.
[37] Y. Lou, C. Zhang, Y. Zheng, X. Xie, W. Wang, and Y. Huang, "Map-matching for low-sampling-rate GPS trajectories," in GIS: Proceedings of the ACM International Symposium on Advances in Geographic Information Systems, 2009, pp. 352–361, doi: 10.1145/1653771.1653820.
[38] S. Brecheisen, H. P. Kriegel, P. Kröger, and M. Pfeifle, "Visually mining through cluster hierarchies," in SIAM Proceedings Series, 2004, pp. 400–411, doi: 10.1137/1.9781611972740.37.
[39] M. Oguchi, T., Katakura, M., & Taniguchi, "Available concepts of energy reduction measures against road vehicular traffic," 1996.
[40] B. G. ROBERT, "BOSCH AUTOMOTIVE HANDBOOK - 5TH EDITION," 2001. https://trid.trb.org/view.aspx?id=690675 (accessed Sep. 23, 2020).
[41] S. CV.AB., "Technical report 7013808," 2012.
[42] M. Muratori, J. Holden, M. Lammert, A. Duran, S. Young, and J. Gonder, "Potentials for platooning in U.S. highway freight transport," SAE Int. J. Commer. Veh., vol. 10, no. 1, pp. 45–49, May 2017, doi: 10.4271/2017-01-0086.

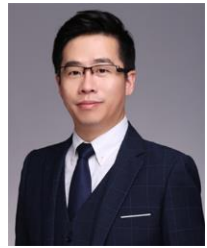

**Xiaolei Ma** is an associate Professor in the school of transportation science and engineering at Beihang University, China. He received his Ph.D. degree from University of Washington, Seattle in 2013. His research areas mainly lie in public transit operation and planning, shared mobility modeling and transportation Big Data analytics. To date, he has published over 100 articles in peer-reviewed journal articles and conference papers. Dr. Ma servers as an associate editor of IEEE Transactions on Intelligent Transportation Systems and IET Intelligent Transport Systems while also being an editorial member of Transportation Research Part C. He is also a member of the TRB Artificial Intelligence and Advanced Computing Applications Committee. Dr. Ma received several academic awards including Young Elite Scientist Sponsorship Program by the China Association for Science and Technology, Beijing Outstanding Youth Program and Beijing Nova Program.

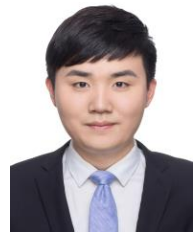

**Enze Huo** is currently pursuing the M.S. degree with the Beijing Key Laboratory for Cooperative Vehicle Infrastructure Systems and Safety Control, School of Transportation Science and Engineering, Beihang University, Beijing, China. His research interests include traffic data mining and freight platoon optimization.

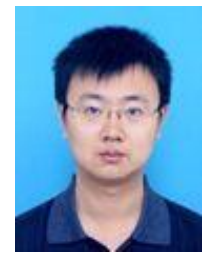

**Haiyang Yu** is an associate professor of the School of Transportation Science and Engineering at Beihang University. Haiyang Yu received his Ph.D. degree in traffic environment and safety technology from Jilin University in 2012. His research interesting includes traffic big data, state characteristic information extraction and expression for road network and traffic control and simulation.

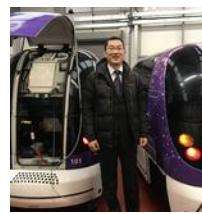

**Honghai Li** is currently pursuing the Ph.D. degree with the Beijing Key Laboratory for Cooperative Vehicle Infrastructure Systems and Safety Control, School of Transportation Science and Engineering, Beihang University, Beijing, China. His research interests include traffic data analytics and intelligent transportation systems.